\documentclass[letterpaper]{article} 
\usepackage{aaai25}  
\usepackage{times}  
\usepackage{helvet}  
\usepackage{courier}  
\usepackage[hyphens]{url}  
\usepackage{graphicx} 
\usepackage{natbib}  
\usepackage{caption} 
\usepackage{algorithm}
\usepackage{algorithmic}

\usepackage{newfloat}
\usepackage{listings}
\DeclareCaptionStyle{ruled}{labelfont=normalfont,labelsep=colon,strut=off} 
\floatstyle{ruled}
\newfloat{listing}{tb}{lst}{}
\floatname{listing}{Listing}
\usepackage{times}
\usepackage{soul}
\usepackage[utf8]{inputenc}
\usepackage{booktabs}
\usepackage{amsmath}
\usepackage{amsthm}
\usepackage{algorithm}
\usepackage{algorithmic}
\usepackage[switch]{lineno}
\usepackage{subcaption}
\usepackage{multirow}
\usepackage{amssymb}
\usepackage{pifont}
\usepackage{tabularx}
\usepackage{mdframed}
\usepackage{xcolor}
\usepackage{xurl}
\usepackage{rotating}
\usepackage{tabularray}
\usepackage{bbding}       
\usepackage{fontawesome} 

\title{Unsupervised Domain Adaptive Person Search via Dual Self-Calibration}
\author {
    Linfeng Qi\textsuperscript{\rm 1},
    Huibing Wang\textsuperscript{\rm 1}\thanks{Corresponding Author.},
    Jiqing Zhang\textsuperscript{\rm 1},
    Jinjia Peng\textsuperscript{\rm 2},
    Yang Wang\textsuperscript{\rm 3}
}
\affiliations {
    \textsuperscript{\rm 1}School of Information Science and Technology, Dalian Maritime University, Dalian, China\\
    \textsuperscript{\rm 2}School of Cyber Security and Computer, Hebei University, Baoding, China\\
    \textsuperscript{\rm 3}School of Computer Science and Information Engineering, Hefei University of Technology, Hefei, China\\
    \{qilinfeng, huibing.wang, jqz\}@dlmu.edu.cn,
    pengjinjia@hbu.edu.cn,
    yangwang@hfut.edu.cn
}

\usepackage{bibentry}

\begin{document}

\maketitle

\begin{abstract}
Unsupervised Domain Adaptive (UDA) person search focuses on employing the model trained on a labeled source domain dataset to a target domain dataset without any additional annotations. Most effective UDA person search methods typically utilize the ground truth of the source domain and pseudo-labels derived from clustering during the training process for domain adaptation. However, the performance of these approaches will be significantly restricted by the disrupting pseudo-labels resulting from inter-domain disparities. In this paper, we propose a Dual Self-Calibration (DSCA) framework for UDA person search that effectively eliminates the interference of noisy pseudo-labels by considering both the image-level and instance-level features perspectives. Specifically, we first present a simple yet effective Perception-Driven Adaptive Filter (PDAF) to adaptively predict a dynamic filter threshold based on input features. This threshold assists in eliminating noisy pseudo-boxes and other background interference, allowing our approach to focus on foreground targets and avoid indiscriminate domain adaptation. Besides, we further propose a Cluster Proxy Representation (CPR) module to enhance the update strategy of cluster representation, which mitigates the pollution of clusters from misidentified instances and effectively streamlines the training process for unlabeled target domains. With the above design, our method can achieve state-of-the-art (SOTA) performance on two benchmark datasets, with 80.2\% mAP and 81.7\% top-1 on the CUHK-SYSU dataset, with 39.9\% mAP and 81.6\% top-1 on the PRW dataset, which is comparable to or even exceeds the performance of some fully supervised methods. Our source code is available at {https://github.com/whbdmu/DSCA}.
\end{abstract}

\section{Introduction}

\begin{figure}[tbp!]
	\centering
	\includegraphics[width=0.97\linewidth]
    {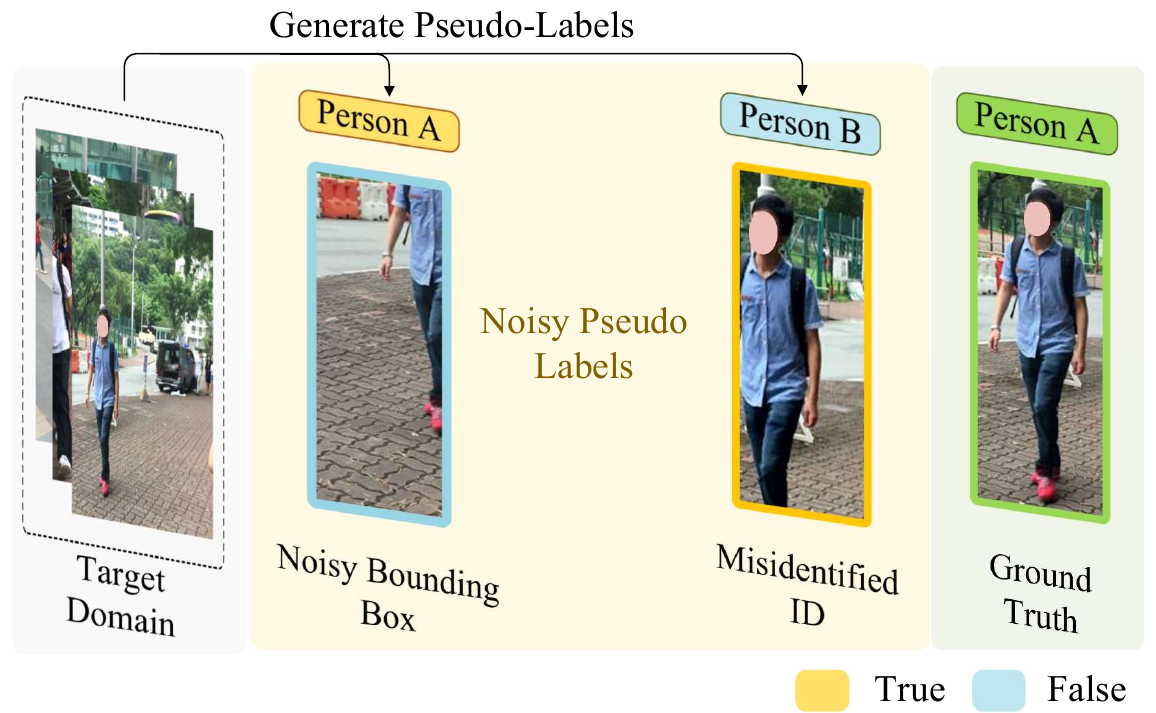}
	\caption{Noisy pseudo-labels consist of low-quality pseudo bounding boxes and misidentified pseudo identities. Where the quality of bounding boxes is the result of unsupervised detection, the reliability of the identities is affected by the clustering features.}
	\label{fig_motivation}

\end{figure}
Person search is a task that combines pedestrian detection~\cite{girshick2014rich,ren2015faster} and re-identification (ReID) \cite{ye2020cross,peng2020unsupervised,wang2022progressive,ye2023channel} to achieve localization and identification of the target pedestrians in real-life scenes. Supervised learning in person search tasks has made significant advancements~\cite{jaffe2023gallery,jiang2024scene}. However, models trained in one domain often struggle to generalize effectively to other domains due to factors such as person viewpoints and camera configuration~\cite{liu2024structure}. Moreover, annotating sufficient training data for a specific scenario is laborious and expensive. Therefore, unsupervised domain adaptation (UDA)  offers significant potential in real-world scenarios for person search.

UDA \cite{ganin2015unsupervised,kang2019contrastive} for person search aims to narrow the gap between ideal training data and the realities of practical applications. However, this field is still in its infancy.  Current approaches involve combining the pre-trained model on the source domain with clustering to produce pseudo labels on the target domain, which are then employed to facilitate the feature representation learning \cite{huang2023self,wang2024manifold}. Specifically, 
DAPS \cite{li2022domain} is the pioneering work that proposes an implicit alignment module to transfer the knowledge of well-annotated source domain data to the unlabeled target domain for UDA person search. Based on this, DDAM \cite{almansoori2024ddam} further enhances domain adaptation by generating a hybrid domain representation, which improves knowledge transfer between the source and target domains. Although effective, these methods fail to acknowledge that pseudo-labels frequently include incorrect labels due to domain gaps. As shown in Figure \ref{fig_motivation}, the incorrect pseudo-labels are comprised of two parts: the noisy pseudo bounding boxes obtained from unsupervised detection, and the misidentified pseudo-IDs obtained from the clustering of the extracted instance-level features. These noisy labels potentially lead to misleading feature learning and impair the effectiveness of domain adaptation.

To tackle the above challenges, we propose a novel framework dubbed Dual Self-Calibration (DSCA) that eliminates the interference caused by noisy pseudo-labels on the UDA person search from both the image and instance perspectives. Specifically, we first present a Perceptually Driven Adaptive Filter (PDAF) to effectively identify and eliminate invalid regions of interest from image-level features, alleviating distractions from incorrect pseudo bounding boxes.  Our PDAF consists of two key components: Perception-Driven Threshold (PDT) and Self-Calibrating Filter (SCF). The SCT has the ability to adaptively filter misjudged and meaningless features according to the threshold predicted by PDT, effectively enhancing the robustness of domain adaptation. Besides, to address the issue of clusters being contaminated by misidentified pseudo-ID instances, we further develop a novel Cluster Proxy Representation (CPR) to enhance the cluster update strategy. The CPR designs a memory structure that moves the utilization and update of the memory dictionary from the instance level to the cluster level. Misidentified instances are treated as part of a cluster proxy rather than as separate instances. As cluster proxies are continuously updated, the effect of misidentified instances will be eliminated. Note that our DSCA only requires clustering once per-training epoch, unlike other methods that generate pseudo-labels through multiple clustering~\cite{ge2020self}. Meanwhile, the memory dictionary merely utilizes and updates the unique cluster proxy for instances in the same cluster. These measures effectively simplified the training process and thus improved efficiency. Extensive experiments on different target domain datasets validate the effectiveness of the proposed DSCA, which outperforms existing SOTA unsupervised domain adaptive methods by significant margins. Ablation experiments also evidence the importance of each key component of DSCA.

In summary, this paper makes the following contributions:
    \begin{itemize}
    \item This paper proposes a novel dual self-calibration framework for UDA person search. The framework ensures the effectiveness of domain adaptation by eliminating the negative effects of potentially erroneous pseudo-labels. Its performance is superior to the current SOTA methods.
    \item We present the Perception-Driven Adaptive Filter to suppress invalid pseudo bounding boxes at different scales in the backbone.
    \item We propose a simplified and efficient cluster update strategy, Cluster Proxy Representation, to address the contamination of clusters by misidentified instances.
    \end{itemize}
\section{Method}

\begin{figure*}[t!]
	\centering
	\includegraphics[width=0.97\linewidth]{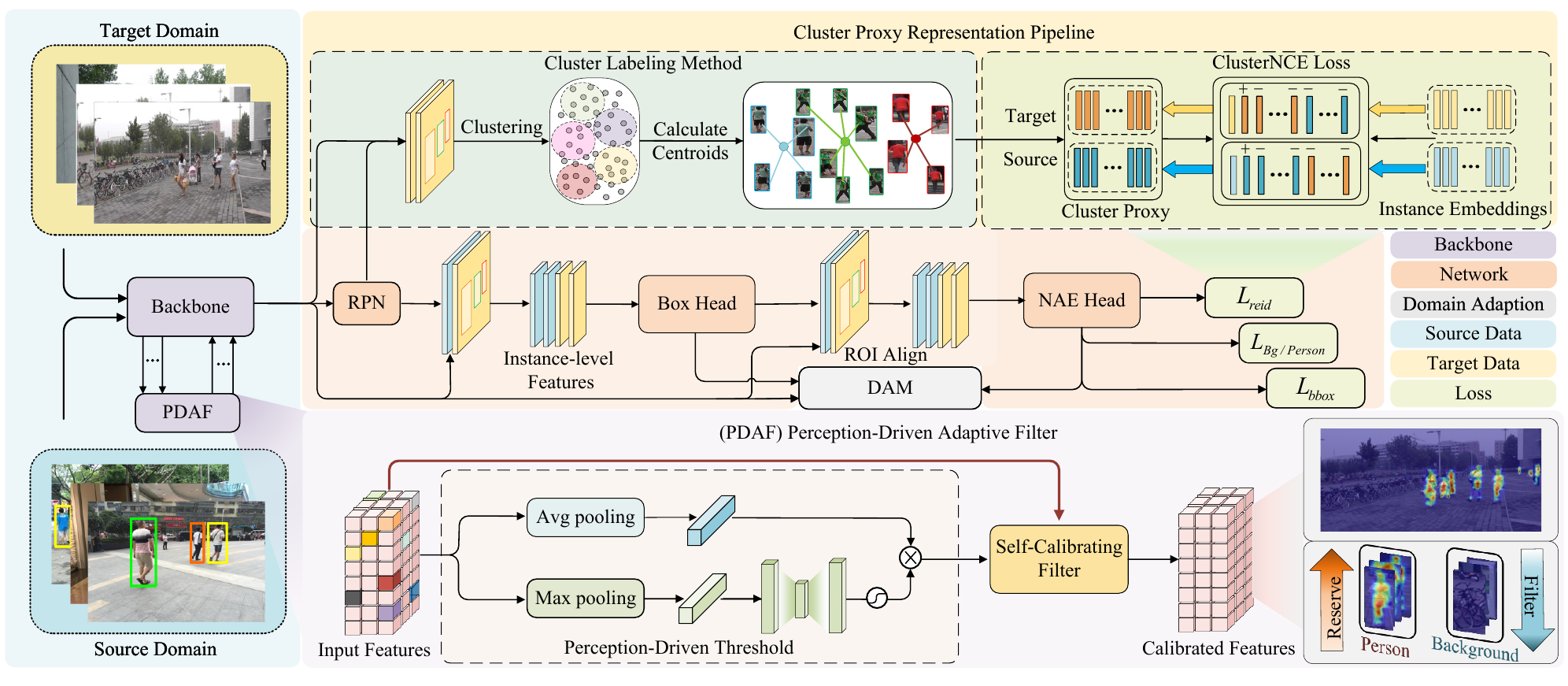}
	\caption{The design architecture of the DSCA framework. For each training period, DSCA alternates between two phases:(1) Cluster Proxy Representation Pipeline. Using RPN generated proposal boxes to annotate unlabeled samples in the target domain, the annotated samples are clustered to assign pseudo-labels and initialize the cluster proxy dictionary. (2) Perception-Driven Adaptive Filter. Image-level features with valid foreground information after PDAF purification are employed for downstream domain alignment and person search tasks.}
	\label{f2}

\end{figure*}

In this section, we introduce the Dual Self-Calibration (DSAC) framework. First, we outline the network architecture. Second, the details of the Perception Driven Adaptive Filter (PDAF) proposed in this paper are elaborated. Finally, we illustrate the working principle of Clustered Proxy Representation (CPR).

\subsection{Framework overview}
The person search model is based on the end-to-end architecture of the DAPS network, which includes an implicit domain alignment module (DAM) \cite{chen2018domain,deng2018image} designed to reduce the discrepancy between two domains. By this approach, we adhere to the principle of ``sample quality first" and utilize our novel method to obtain more refined feature representation and pseudo-labels for the UDA tasks. 

As illustrated in Figure \ref{f2}, the DSCA framework comprises perception-calibrated person search and unlabeled target domain preprocessing. 1) In perception-calibrated person search, DSCA uses the first four convolutional layers of ResNet-50 \cite{he2016deep} as a backbone network to extract image-level features from the input images in both the source and target domains. Implement PDAF for each layer of the backbone network to ensure the concentration of extracted features on the foreground targets, thereby facilitating multi-scale feature filtering. Subsequently, reliable image-level features are fed into the detection network via RPN\cite{ren2015faster} generate region proposals and regressed through the box head to output the detected bounding boxes. Finally, based on the instance-level features extracted from the bounding boxes, the more sophisticated detection and re-identification tasks are continued at the NAE head \cite{chen2020norm}. The DAM is used to eliminate inter-domain disparities at both the image and instance scales. 2) Before each epoch's training phase, DSCA applies the CPR pipeline for unlabeled target domain preprocessing. The network directly extracts training proposals from the target domain using RPN. The boxes obtained from training proposals are directly utilized as pseudo bounding boxes. To mitigate the impact of misidentified instances, DSCA presents CPR for training proposals, employing a single clustering strategy that directly assigns and utilizes pseudo-IDs. Furthermore, the average features within clusters are applied to initialize the memory dictionary for Contrastive Learning \cite{he2020momentum,dai2022cluster}. 

\subsection{Perception-Driven Adaptive Filter}
This paper proposes a Perception-Driven Adaptive Filter (PDAF). Through multi-scale filtering in the backbone, it eliminates misfocusing on scene information and ensures task flow reliability.
\subsubsection{Perception-Driven Threshold.} 
Backbone network is a powerful feature extractor designed to provide effective feature representations for downstream tasks. However, during the learning process of UDA, some invalid feature representations are likely to be erroneously recognized as targets and used for generating pseudo-labels due to their highlighting. Therefore, to obtain accurate filtering thresholds to eliminate these invalid representations, this paper proposes a Perception-Driven Threshold (PDT). The PDT distinguishes scene information based on the foreground target perception capability obtained by training the model on the source domain.

Specifically, PDT generates a channel attention map using the inter-channel relationships of the features and uses this as an adaptive threshold. To achieve this, we used pooling to compress the input image-level features' dimensions for spatial information aggregation. With the average pooling and max pooling operations, we aggregate the spatial information of image-level features $F\in {R}^{H\times W\times C}$ from different perspectives to obtain two spatial contextual representations: $F_{avg}\in {R}^{1\times 1\times C}$ and $F_{max}\in {R}^{1\times 1\times C}$. 

In the context of person search, average pooling aggregates global spatial messages and evaluates the overall effectiveness of the channel. Therefore, in this paper, we consider $F_{avg}\in {R}^{1\times 1\times C}$ as a kind of basic threshold  $\tau^{\prime}\in {R}^{1\times 1\times C}$ and use it as a benchmark for distinguishing between foreground and background information. It can be described as:
\begin{equation}
    \tau^{\prime} = AvgPool\left(F\right) ,
\end{equation}
while $\tau^{\prime}$ achieves a basic ability to distinguish and shows good performance in fixed scenarios, it is not flexible enough. For complex real-world scenarios, its effectiveness will be greatly reduced. Therefore, PDT uses the max pooling features $F_{max}$ to guide $\tau^{\prime}$ to ensure that the obtained thresholds can be adapted to scene changes. This is because max pooling features $F_{max}$ can be employed as a key indicator to ascertain the presence of foreground information in channels, and they are relatively less sensitive to scene variations \cite{wei2021fine}. Specifically, $F_{max}$ be passed into a multi-layer perceptron (MLP) and use the sigmoid function to scale the output to the range of $0$ to $1$, so we can obtain the adaptive scaling factor. Finally, we use element-wise multiplication to obtain the PDT. In short, it can be calculated as:
\begin{equation}
    \begin{split}
    \alpha & = \sigma \left (MLP\left (MaxPool\left(F\right)\right)\right) ,\\
    \tau & = \alpha \left(AvgPool\left(F\right)\right) ,
    \end{split}
\end{equation}
where $\sigma$ denotes the sigmoid function, $\alpha\in {R}^{1\times 1\times C}$ represents the adaptive scaling factor, and $\tau\in {R}^{1\times 1\times C}$ represents our Perception-Driven Threshold. 

\subsubsection{Self-Calibrating Filter.}
The design of the threshold function is of paramount importance in the context of filter design. Traditional threshold functions \cite{donoho1994ideal,donoho1995noising} encompass both hard and soft thresholds. However, neither of these functions is optimal for person search. 

The hard threshold function can directly filter out the part of the input features that are smaller than the threshold value without interfering with the foreground features. It can be expressed as follows:
\begin{equation}
f_{hard}(x,\tau)=
    \begin{cases}
    x & \left | x \right | \ge \tau \\
    0 & \left | x \right |< \tau ,\\
\end{cases} 
\end{equation}
where $x$ denotes the value of the input and $\tau$ means the threshold. However, this simply truncated approach impairs the structural information and affects the ability to model fitting. The soft threshold function performs a panning operation on the input features according to the threshold value and then eliminates the portion below the threshold. It can be described as:
\begin{equation}
f_{soft}(x,\tau)=
    \begin{cases}
    x-\tau &  x  \ge \tau \\
    0 & \left | x \right |<\tau \\
    x + \tau &  x  \le -\tau .
\end{cases} 
\end{equation}

The application of a soft threshold ensures the smooth filtering of features. Nonetheless, the filtering process results in a constant bias between the input and output features, which in turn affects the quality of the foreground features. 

\begin{figure}[!tbp]
	\centering
	\includegraphics[width=0.8\linewidth]{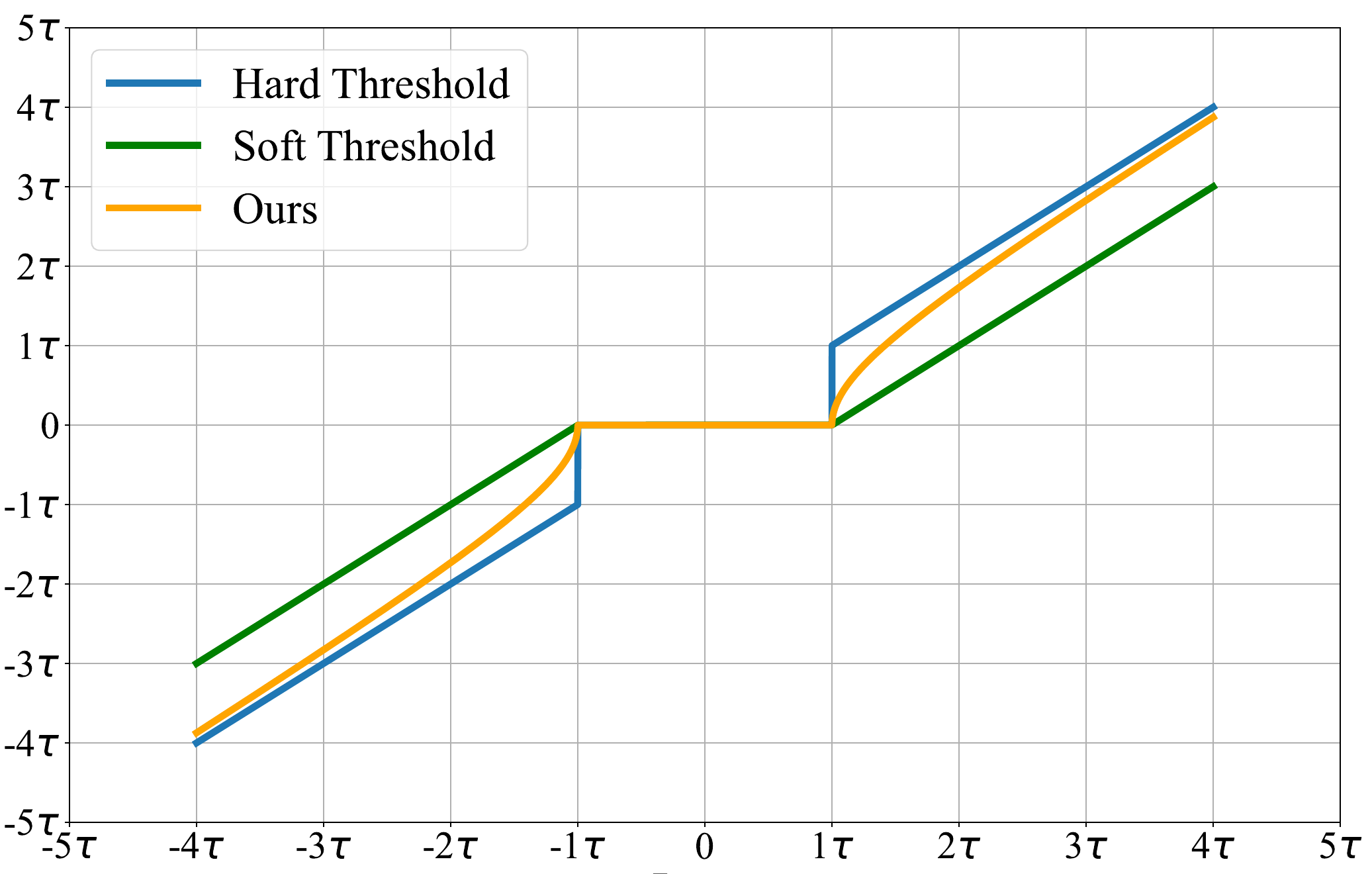}
	\caption{Comparison of our High-Order Soft Threshold with two classical threshold functions, take $n=2$ as an example.}
	\label{f3}

\end{figure}

Based on the above analyses, we rethink the design of the threshold function and propose a Self-Calibrating Filter (SCF). It fully combines the advantages of the traditional threshold functions while overcoming the shortcomings of both. This filter consists of a High-Order Soft Threshold  (HST) and a Self-Calibrating Transition (SCT) in two parts. Specifically, HST further improves the accuracy of the filter by using a high-order function instead of the traditional linear method. As shown in Figure \ref{f3}, HST fully adapts to the needs of the person search task and grasps the differences between the foreground features and scene features. The formula of HST is as follows:
\begin{equation}
f_{hst}(x,\tau)=
    \begin{cases}
    0 & |x|<\tau \\
    sgn(x){\left({{|x|}^n-{\tau}^n}\right)}^{\frac{1}{n}} &  |x|  \ge \tau ,\\
    \end{cases} 
\end{equation}
where $sgn()$ returns the sign of the input value. Although theoretically, the larger $n$ is, the better the function fitted by HST. In actual applications, due to the use of higher-order functions instead of linear computation, neural networks may suffer from potential gradient explosion and gradient vanishing problems. This makes it difficult for gradient descent-based optimization algorithms to update model parameters effectively. To solve this problem, we propose the Self-Calibrating Transition (SCT) strategy. By introducing learnable parameters, we can self-calibrate and adjust the form of the threshold function. This allows the function to be closer to the soft threshold at the beginning of training, and then gradually transition to HST as the training proceeds. The formula of SCF is as follows:

\begin{equation}
\begin{aligned}
    f_{scf}(x,\tau)=
   \text{\scriptsize $
       \begin{cases}
       \lambda \left({{x}^2-{\tau}^2}\right)^{\frac{1}{2}} + \left(1-\lambda\right) \left(x-\tau\right) &  x \ge \tau \\
       0 & \left | x \right |<\tau \\
       -\lambda \left({{x}^2-{\tau}^2}\right)^{\frac{1}{2}}  + \left(1-\lambda\right) \left(x+\tau\right)&  x  \le -\tau ,
       \end{cases} $}
   \end{aligned}
\end{equation}

where $\lambda \in {R}^{1\times 1\times C}$ is a set of learnable parameters initialized as $0$. This gradual transition strategy can not only effectively alleviate the gradient problem but also give full play to the advantages of HST in terms of expressive ability.

\subsection{Cluster Proxy Representation}
To reduce the difficulty of pseudo-labels optimization in UDA person search, some previous studies have designed an instance-level contrastive learning framework inspired by SPCL \cite{ge2020self}. Specifically, a memory dictionary containing all instances is built and updated, but the utility is at the cluster level. This approach can lead to meaningless or even misidentified samples being included as target instances in the training process.

In this paper, we design the Cluster Proxy Representation (CPR). CPR is different from existing memory structures. It uses a unique proxy to represent each category during training, and it updates the memory by cluster, thus effectively reducing the effect of instance pollution.

\subsubsection{Cluster Labeling Method.}
This paper uses the person search framework pre-trained in the source domain to generate training proposals and pseudo bounding boxes, then uses the DBSCAN \cite{ester1996density} algorithm to assign the pseudo-labels. After the above operation, the unlabeled dataset $D_T$ becomes $D_{T}^\prime=\left \{(x_i, l_i^1,...,l_i^n) \right \}_{i=1}^N$, where $x_i$ denotes the $i$-th image in this dataset, $l_i$ is the pseudo-label of the instances in this image, $N$ denotes the number of images, and $n$ denotes the number of instances in the image.

For each cluster, whether it belongs to the source or target domain, we maintain a unique cluster proxy representation, which is initialized with the average instance features of it:
\begin{equation}
	c_k=\frac{1}{N_k} \sum_{f_i\in F_k}f_i ,
	\label{cc}
\end{equation}
where $c_k$ denotes the cluster proxy representation for the $k$-th cluster, $F_k$ represents a set of features with the same cluster $k$,  and $f_i$ is the feature of each instance. Then use the memory dictionary $M=\left \{(c_1, c_2,...,c_K) \right \}$ to store these representation uniformly. 

\begin{table*}[t]
    \small      
    \centering
    \begin{tabularx}{1\linewidth}{
            >{\hsize=.4\hsize \linewidth=\hsize \raggedright\arraybackslash}X
            |>{\hsize=.18\hsize \linewidth=\hsize \raggedright\arraybackslash}X
            *{2}{|*{2}{>{\hsize=.07\hsize \linewidth=\hsize \centering\arraybackslash}X}}
        }
        \toprule
        \multicolumn{1}{c|}{} &
        \multicolumn{1}{c|}{} &
        \multicolumn{2}{c|}{\textbf{PRW}} &
        \multicolumn{2}{c}{\textbf{CUHK-SYSU}} \\
        \cline{3-6}
        \multicolumn{1}{c|}{\multirow{-2}{*}{\textbf{Method}}} &
        \multicolumn{1}{c|}{\multirow{-2}{*}{\textbf{Backbone}}} &
        \multicolumn{1}{c}{mAP} &
        \multicolumn{1}{c|}{top-1} &
        \multicolumn{1}{c}{mAP} &
        \multicolumn{1}{c}{top-1} \\
        \midrule
        \multicolumn{6}{c}{\textit{Fully-Supervised}} \\
        OIM \cite{xiao2017joint} & ResNet-50   & 21.3 & 49.4 & 75.5 & 78.7 \\
        MGTS\cite{chen2018person}   & VGG-16   & 32.6 & 72.1 & 83.0  & 83.7     \\
        HOIM\cite{chen2020hierarchical} & ResNet-50 & 39.8 & 80.4 & 89.7 & 90.8 \\
        NAE+ \cite{chen2020norm} & ResNet-50 & 44.0 & 81.1 & 92.1 & 92.9 \\
        RDLR \cite{han2019re} & ResNet-50 & 42.9 & 70.2 & 93.0 & 94.2  \\
        TCTS \cite{wang2020tcts} & ResNet-50 & 46.8 & 87.5 & 93.9 & 95.1  \\
        AlignPS+ \cite{yan2021anchor} & ResNet-50  & 46.1 & 82.1 & 94.0 & 94.5\\
        SeqNet \cite{li2021sequential} & ResNet-50  & 46.7 & 83.4 & 93.8 & 94.6\\
        PSTR \cite{cao2022pstr} & PVTv2-B2 & 56.5 & 89.7 & 95.2 & 96.2 \\
        SeqNeXt \cite{jaffe2023gallery} & ConvNeXt-B & 57.6 & 89.5 & 96.1 & 96.5 \\
        SEAS\cite{jiang2024scene} & ConvNeXt-B & 60.5 & 89.5 & 97.1 & 97.8 \\
        \midrule
        \multicolumn{6}{c}{\textit{Weakly-Supervised}} \\
        CGPS\cite{yan2022exploring} & ResNet-50 & 16.2 & 68.0 & 80.0 & 82.3 \\
        R-SiamNet\cite{han2021weakly} & ResNet-50  & 21.2 & 73.4 & 86.0 & 87.1\\
        SSL\cite{wang2023self} & ResNet-50 & 30.7 & 80.6 & 87.4 & 88.5 \\
        DICL\cite{wang2024deep} & ResNet-50 & 35.5 & 80.9 & 87.4 & 88.8 \\
        \midrule
        \multicolumn{6}{c}{\textit{Unsupervised}} \\
        DAPS\cite{li2022domain} & ResNet-50  & 34.7  & 80.6 & 77.6  & 79.6   \\
        DDAM\cite{almansoori2024ddam}  & ResNet-50  & 36.7  & 81.2 & 79.5  & 81.3   \\
        \textbf{DSCA(Ours)} & ResNet-50  & \textbf{39.9} & \textbf{81.6} & \textbf{80.2} & \textbf{81.7} \\
        \bottomrule
    \end{tabularx}
    \caption{Comparison of mAP($\%$) and top-1 accuracy($\%$) with the fully supervised, weakly supervised, and unsupervised methods on CUHK-SYSU and PRW datasets.}
    \label{tab:sota}

\end{table*}

\subsubsection{Cluster Memory Updating.}
For CPR, memory updates consist of two parts: online and offline. For the online phase, this work recognizes the $M$ as a look-up table (LUT) \cite{xiao2017joint} that stores all marked pedestrian proxy representations and calculates the loss using the following equation:
\begin{equation}
	L= -\log \frac{\exp \left( f\cdot c_+/\tau \right)}{ {\textstyle \sum_{k=0}^{K}}\exp \left( f\cdot c_k/\tau\right) },
	\label{l1}
\end{equation}
where $f$ is an instance query feature extracted for re-identification and $c_+$ shares the same cluster with $f$. $\tau$ is a temperature factor that adjusts the softness of the probability distribution.  During backward, if the target label is $k$, then we will update the $k$-th column of the $M$ by 
\begin{equation}
	c_k\longleftarrow \left( 1-\gamma \right) c_k+\gamma f ,
	\label{update_c_k}
\end{equation}
where $\gamma \in [0,1]$  is the momentum factor used to control the update speed of the proxy.  Through this cluster-level update strategy, the pollution caused by misidentified instances will be overwritten by subsequent updates.

In the offline stage, to confirm the effectiveness of the cluster proxy in the entire UDA learning process, this paper uses an asynchronous learning strategy. In each training epoch, the clustered proxy and pseudo-labels are reinitialized together. At the same time, to ensure consistency of the training objectives, we introduce instance features extracted from the previous epoch and smooth the instances using the Exponential Moving Average (EMA). This process can be described as follows: 
\begin{small}
\begin{equation}
    \centering
	c_k=\frac{1}{N_k} \sum_{f_i^{t}\in F_k^{t}} \left[ m\left(Match\left(F^{t-1},f_i^{t}\right)\right)+\left(1-m\right)f_i^{t}\right],
	\label{update_offline}
\end{equation}
\end{small}
 where $t$ and $t-1$ represent the training epoch.  $F^{t-1}$ represents all the instance features extracted by the previous epoch. The $Match$ operation refers to returning the instance in  $F^{t-1}$ that coincides with the pseudo bounding box corresponding to the current instance $f_i^{t}$, and $m$ represents the smoothing factor. 
\section{Experiments}

\subsection{Experimental Settings}
\subsubsection{Datasets.} We evaluate our DCSA on two person search datasets: CUHK-SYSU \cite{xiao2017joint} and PRW \cite{zheng2017person}. The CUHK-SYSU  dataset is a large-scale person search benchmark comprising a total of 18,184 images. The dataset contains 8,432 unique person identities and 96,143 annotated bounding boxes. It is divided into a training set with 5,532 identities and 11,206 images, and a test set with 2,900 query persons and 6,978 gallery images. The PRW dataset includes 11,816 scene images with annotations for 932 unique person identities and 43,110 bounding boxes. The training set comprises 932 identities with 5,704 images, while the test set contains 2,057 query persons and 6,112 scene images.

\subsubsection{Metrics.} We adopt two widely used metrics for quantitatively evaluating person search performance: the mean Average Precision (mAP) and top-1 accuracy. Note that all evaluations are performed on the test set of the target domain without any additional annotations.

\subsubsection{Implementation Details.} We implemented our DSCA using Pytorch and trained it on an NVIDIA A800 GPU with a batch size of 4. We adopted the Stochastic Gradient Descent (SGD) optimizer with a learning rate of 0.0024, which is warmed up in the first epoch. We employed the pre-trained ResNet50  \cite{he2016deep} as our backbone. During training, the input images are resized to 1500$\times$900, and random horizontal flipping is applied for data augmentation. We set both the momentum factor $\gamma$ and smoothing factor $m$ to 0.2 for online and offline cluster updating, respectively. To prevent overfitting, we adjust the number of epochs based on the variations in the target domain dataset. Specifically, when PRW is used as the target domain, our DSCA undergoes pre-training on the source domain CUHK-SYSU for 7 epochs before commencing joint training for 13 epochs. Conversely, we first pre-train on the source domain PRW for 2 epochs, followed by beginning joint training for 7 epochs.

\subsection{Comparison with State-of-the-Art Methods}
\begin{figure}[!t]
    \centering
    \includegraphics[width=0.95\linewidth]{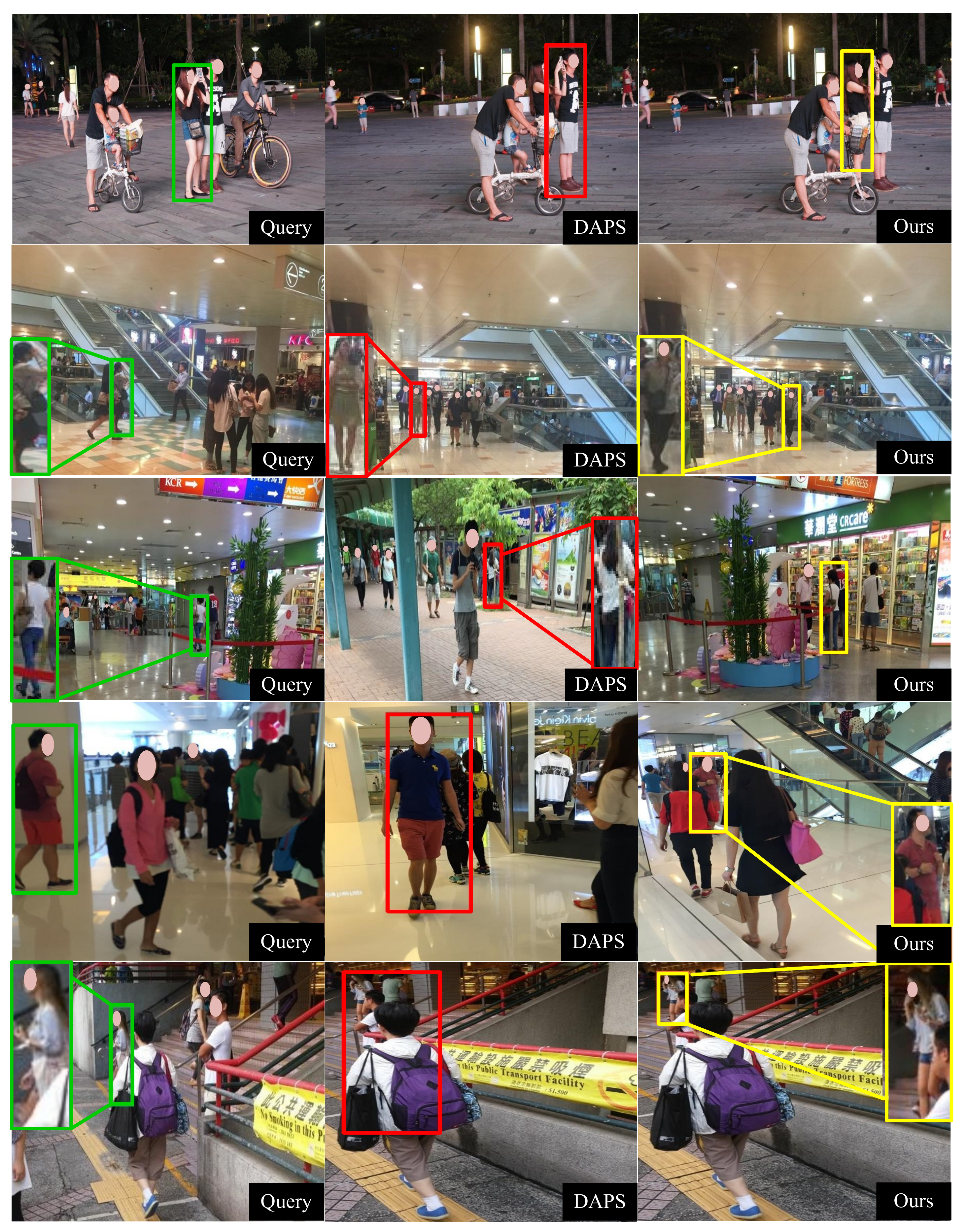}
    \caption{Qualitative comparison of DSCA with DAPS on the CUHK-SYSU test set. The green bounding boxes denote the queries, while the red and orange bounding boxes denote incorrect and correct matches, respectively.}
    \label{fig:vis}

\end{figure}

To demonstrate the effectiveness of our method, we compare DSCA with SOTA approaches that employ various training strategies, including fully supervised, weakly supervised, and unsupervised methods. As shown in Table~\ref{tab:sota}, our DCSA outperforms other compared unsupervised domain adaptive approaches on both two datasets in terms of both mAP and top-1 scores. For example, our method achieves 80.2\% overall mAP and 81.7\% top-1 on the CUHK-SYSU dataset, outperforming the runner-up by 0.7\% and 0.4\%, respectively. This demonstrates that eliminating the disruptive influence of noisy pseudo-labels in domain adaptation can contribute to enhancing the robustness of UDA person search. Figure~\ref{fig:vis}  further qualitatively demonstrates the benefits of our method in some challenging conditions, including cross-scene, low-resolution, occlusion, and viewpoint variation. 

\begin{figure}[t]
	\centering
	\includegraphics[width=0.8\linewidth]{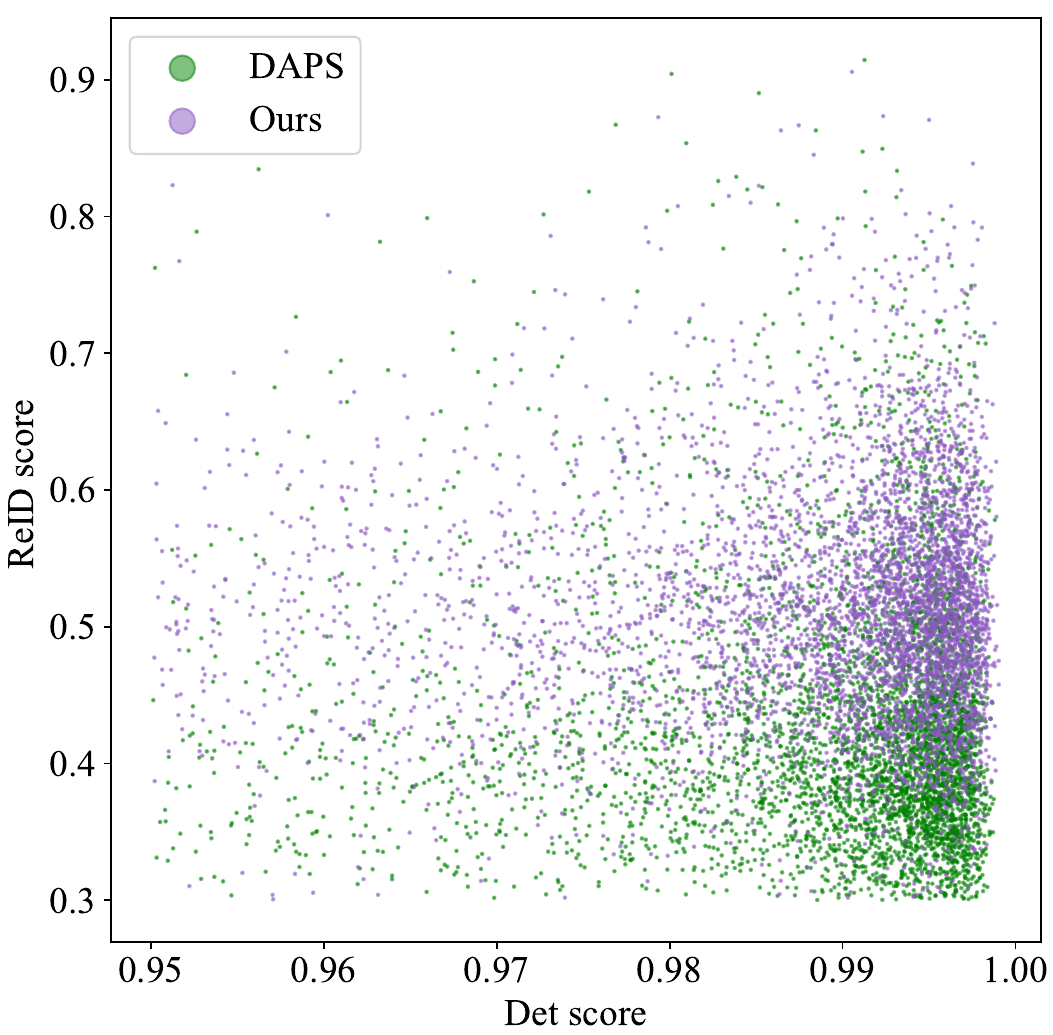}
	\caption{Visualization of the detection scores and re-identification scores on the CUHK-SYSU dataset. Purple and green dots indicate the results of DAPS and ours.}
	\label{fig_scatter}
\end{figure}

\begin{table}[t]
    \small
    \centering
    \begin{tabularx}{\linewidth}{
            >{\hsize=0.125\hsize \linewidth=\hsize \centering\arraybackslash}X
            *{2}{>{\hsize=.125\hsize \linewidth=\hsize \centering\arraybackslash}X}
            *{2}{|*{2}{>{\hsize=.125\hsize \linewidth=\hsize \centering\arraybackslash}X}}
        }
        \toprule
        \multicolumn{3}{c|}{\textbf{Components}} &
        \multicolumn{2}{c|}{\textbf{PRW}} &
        \multicolumn{2}{c}{\textbf{CUHK-SYSU}} \\
        \cmidrule(r){1-7}
        \multicolumn{1}{c}{PDT} &
        \multicolumn{1}{c}{SCF} &
        \multicolumn{1}{c|}{CPR} &
        \multicolumn{1}{c}{mAP} &
        \multicolumn{1}{c|}{top-1} &
        \multicolumn{1}{c}{mAP} &
        \multicolumn{1}{c}{top-1} \\
        \midrule
        \ding{55} &           &           & 39.0 & 80.8 & 76.9 & 78.3 \\
                  & \ding{55} &           & 38.4 & 80.8 & 79.6 & 81.2 \\
        \ding{55} & \ding{55} &           & 36.2 & 79.7 & 78.2 & 80.5 \\
                   &          & \ding{55} & 39.2 & 81.1 & 79.5 & 80.7 \\
        \ding{55} & \ding{55} & \ding{55} & 34,7 & 80.6 & 77.6 & 79.6 \\
                  &           &           & \textbf{39.9} & \textbf{81.6} & \textbf{80.2} & \textbf{81.7} \\

        \bottomrule
    \end{tabularx}
    \caption{Validating the effectiveness of different components on the CUHK-SYSU and PRW datasets. CPR: Cluster Proxy Representation. PDT: Perception-Driven Threshold. SCF: Self-Calibrating Filter.}
    \label{tab:ablation}

\end{table}

Surprisingly, Table~\ref{tab:sota} shows that our DSCA continues to deliver comparable performance in comparison to both fully supervised and weakly supervised approaches. Particularly on the PRW dataset, our method surpasses all compared weakly supervised methods by a large margin, outperforming the best one by 4.4\% and 0.7\% in mAP and top-1, respectively. Besides, the results show a noticeable disparity between DSCA and SOTA fully supervised methods, suggesting that there is still potential for improvement to bridge this gap. In this work, we prioritize minimizing data dependencies and achieving a favorable balance between performance and resource efficiency. We also anticipate that our approach will stimulate further investigation in this field.

\subsection{Ablation Study }

\begin{figure}[t]
    \centering
    \includegraphics[width=0.8\linewidth]{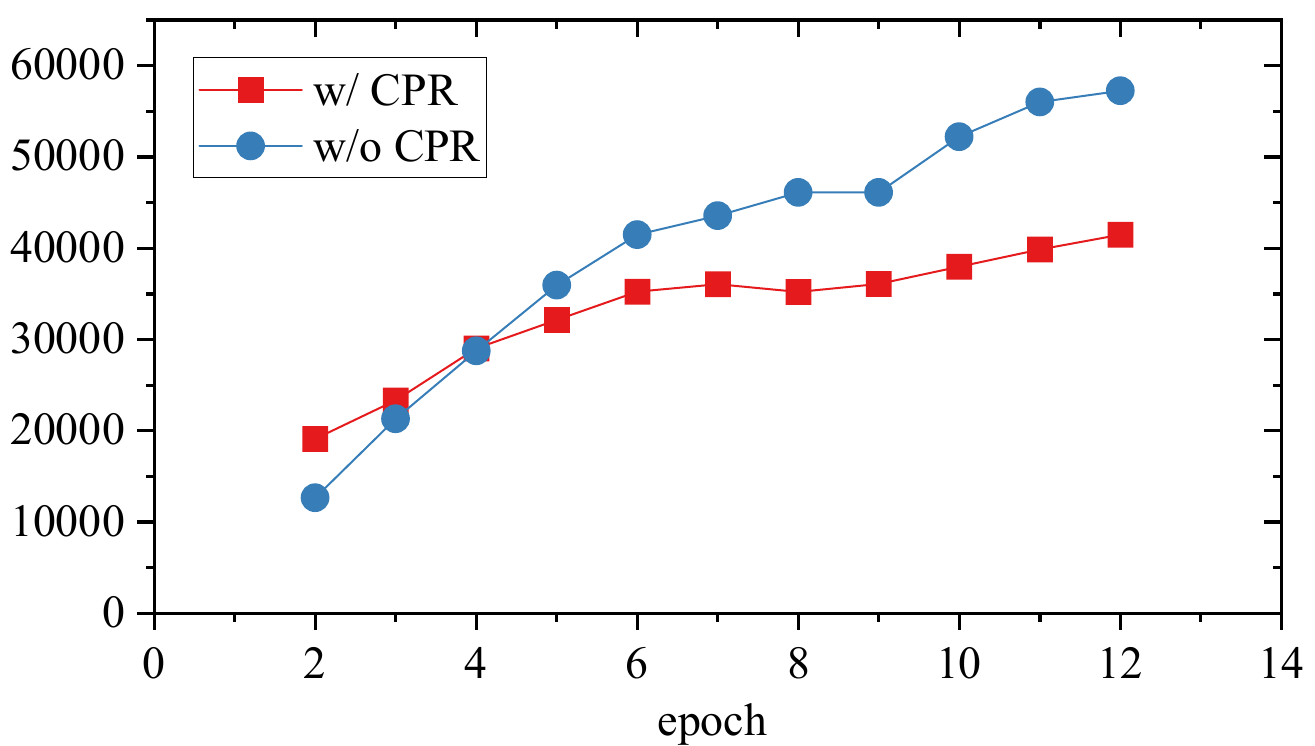}
    \caption{Number of cluster representations stored in the memory dictionary, when the CUHK-SYSU dataset is used as the target domain.}
    \label{fig:vis_cpr}
\end{figure}

\begin{table}[t]
    \small
    \centering
    \begin{tabularx}{\linewidth}{
            >{\hsize=0.125\hsize \linewidth=\hsize \centering\arraybackslash}X
            *{1}{>{\hsize=.125\hsize \linewidth=\hsize \centering\arraybackslash}X}
            *{2}{|*{2}{>{\hsize=.125\hsize \linewidth=\hsize \centering\arraybackslash}X}}
        }
        \toprule
        \multicolumn{2}{c|}{\textbf{PDAF}} &
        \multicolumn{2}{c|}{\textbf{PRW}} &
        \multicolumn{2}{c}{\textbf{CUHK-SYSU}} \\
         \cmidrule(r){1-6}
        \multicolumn{1}{c}{Threshold} &
        \multicolumn{1}{c|}{Filter} &
        \multicolumn{1}{c}{mAP} &
        \multicolumn{1}{c|}{top-1} &
        \multicolumn{1}{c}{mAP} &
        \multicolumn{1}{c}{top-1} \\
        \midrule
        AMT & Hard  & 38.6 & 81.0 & 78.7 & 80.1 \\
        AMT & Soft  & 38.4 & 80.3 & 77.9 & 80.1 \\
        AMT & SCF  & 39.2 & 81.3 & 79.5 & 81.2 \\
        PDT & Hard  & 38.1 & 79.9 & 79.3 & 80.9 \\
        PDT & Soft & 38.4 & 80.8 & 79.6 & 81.2 \\
        PDT & SCF  & \textbf{39.9} & \textbf{81.6} & \textbf{80.2} & \textbf{81.7} \\

        \bottomrule
     \end{tabularx}
    \caption{ Results of the ablation study on Perception-Driven Adaptive Filter, compare results of the different thresholds and filters.}
    \label{tab:ablation_pdaf}

\end{table}

\begin{figure}[t]
    \centering
    \includegraphics[width=\linewidth]{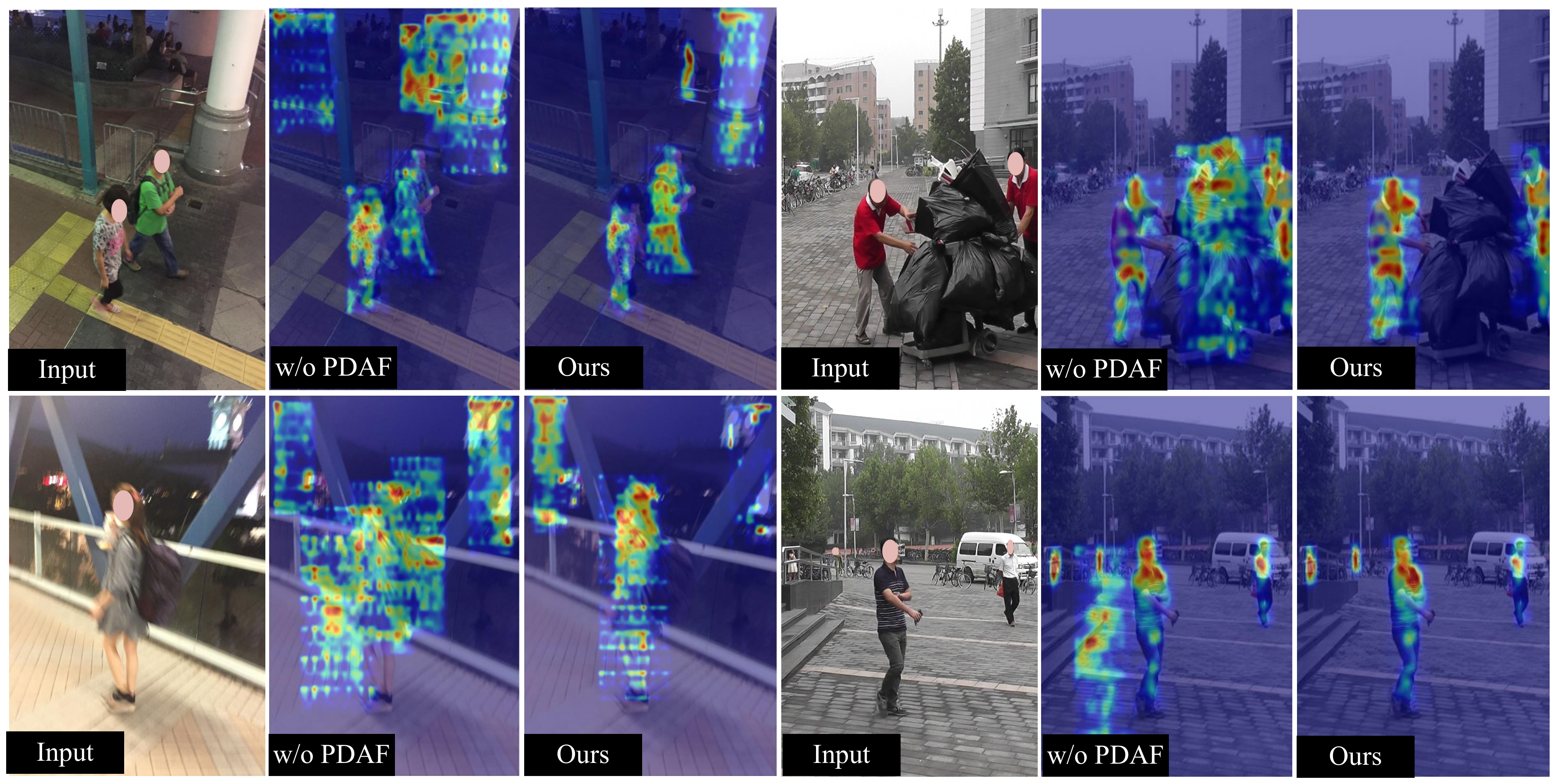}
    \caption{Visualization of the LayerCAM for the last layer of the backbone for each predicted object (images from CUHK-SYSU and PRW datasets). Evaluated separately with and without PDAF, the perceptual scores are differentiated by different colors, from blue to red, representing increasing perceptual scores.}
    \label{fig:vis_pdaf}
\end{figure}

\textbf{Impact on Perception-Driven Adaptive Filter.} 
Our proposed Perception-Driven Adaptive Filter (PDAF) aims to filter out the interference of erroneous pseudo-boxes on domain adaptation, which has two key components: Perception-Driven Threshold (PDT) module and  Self-Calibrating Filter (SCF). We thus conduct the following experiments to validate the effectiveness of PDAF: (i) without PDT (\textit{i.e.}, replacing PDT with the global average operation for input features) (ii) without SCF; (iii) without PDT and SCF. As shown in Table~\ref{tab:ablation} \#1 and \#2, removing either PDT or SCF results in significant performance degradation on both datasets. This validates that the exclusion of noisy pseudo-boxes is essential for person search accuracy. Removing both PDT and SCF leads to an additional decrease in performance, as shown in \#3, providing further evidence of the effectiveness of our design. 

To further validate the effectiveness of PDT and SCF, we conduct additional ablation experiments: (i) replacing PDT with the Adaptive Mean Threshold (AMT) 
 module~\cite{zhao2019deep}, which is a typical structure that mainly employs global average pooling to predict thresholds; (ii) replacing SCF with Hard and Soft Threshold schemes. As shown in Table \ref{tab:ablation_pdaf}, our proposed PDT consistently outperforms AMT counterparts in terms of both mAP and top-1 scores, suggesting the effectiveness of our PDT strategy to combine local foreground and global spatial information. Moreover, Table \ref{tab:ablation_pdaf} shows that the performance of our proposed filtering method SCF considerably exceeds the standard Hard and Soft threshold approaches. This is because conventional soft and hard thresholds are constrained by inter-domain disparities. Specifically, for the scene-rich SYSU-CUHK dataset, the hard threshold cannot adapt to scene variations. In contrast, soft thresholds are inhibited by monotonous scenes in the PRW dataset. Our SCF fully integrates the strengths of both to achieve the best overall performance.

To get more insights into PDAF, we adopt LayerCAM~\cite{jiang2021layercam} to visualize the attention maps of the backbone outputs. As shown in Figure \ref{fig:vis_pdaf}, under the guidance of PDAF, our method pays more attention to the location of the target, greatly reducing noise interference in the domain adaptation process and thus enhancing the reliability of person search tasks.

\subsubsection{Effectiveness of Cluster Proxy Representation.} 
The Cluster Proxy Representation (CPR) scheme is another essential component of our approach, with a focus on enhancing training efficiency by reducing the impact of misidentified instances on cluster pollution. As shown in Table~\ref{tab:ablation} \#4, the removal of CPR results in a decrease in both mAP and top-1 scores on the PRW and CUHK-SYSU datasets, which serves as evidence of the effectiveness of our proposed cluster proxy. As shown in Figure \ref{fig:vis_cpr}, The use of CPR also greatly reduces GPU memory consumption.

\begin{table}[t]
    \small
    \centering
    \begin{tabularx}{\linewidth}{
            >{\centering\arraybackslash}X
            *{1}{|*{3}{>{\centering\arraybackslash}X}}
        }
        \toprule
        \multicolumn{1}{c|}{\textbf{Method}} &
        \multicolumn{1}{c}{$\textbf{Prop}_{0.3}\uparrow$} &
        \multicolumn{1}{c}{$\textbf{Prop}_{0.5}\uparrow$} &
        \multicolumn{1}{c}{$\textbf{Prop}_{0.7}\uparrow$} \\

        \midrule
        Ours & \textbf{99.7\%} & \textbf{52.8\%} & \textbf{2.3\%}  \\
        DAPS & 97.0\% & 17.5\% & 1.8\%  \\
        \midrule
        \multicolumn{1}{c|}{$\Delta(drop)$} &
        \multicolumn{1}{c}{2.7\%} &
        \multicolumn{1}{c}{35.3\%} &
        \multicolumn{1}{c}{0.5\%} \\
        \bottomrule
    
    \end{tabularx}
    \caption{Results obtained from the evaluation of the CUHK-SYSU dataset. $\textbf{Prop}_{0.3}$, $\textbf{Prop}_{0.5}$ and $\textbf{Prop}_{0.7}$ denote the proportion of re-identification scores greater than $0.3$, $0.5$, $0.7$, respectively.}
    \label{tab:ablation_cpr}

\end{table}

\subsubsection{Performance on Person Detection and ReID.} 
To further illustrate the effectiveness of our DSCA, we examine the performance of the two sub-tasks involved in person search: pedestrian detection and ReID. As shown in Figure~\ref{fig_scatter}, our method obtains significantly higher overall scores in both detection and ReID on the CUHK-SYSU dataset when contrasted with another SOTA unsupervised method DAPS. We conduct a more detailed examination of the performance of ReID at various thresholds. Table~\ref{tab:ablation_cpr} shows our method significantly outperforms the DAPS across all evaluated thresholds of ReID scores. Specifically, with a threshold of 0.5, our DSCA achieves a significant rate of 52.8\%, surpassing DAPS by a notable margin of 35.3\%. 
The results demonstrate that our design for mitigating the impact of noisy pseudo-labels significantly enhances the overall performance of both the detection and ReID sub-tasks.

\section{Conclusion}
In this paper, we propose a Dual Self-Calibration (DSCA) framework to remove the interference caused by noisy pseudo-labels due to inter-domain disparities for UDA person search. This disturbance originates from both images and instances. For these two kinds of interference, we design perception-driven adaptive filter and cluster proxy representation to achieve the filtering of noisy pseudo-boxes and the elimination of misidentified instances, respectively.

\bibliography{aaai25}

\end{document}